\documentclass[conference]{IEEEtran}
\IEEEoverridecommandlockouts
\usepackage{algorithm,algpseudocode}
\usepackage{amsmath,amssymb,amsthm,amsfonts}
\usepackage{cite}
\usepackage{graphicx}
\usepackage[hidelinks]{hyperref}

\def\BibTeX{{\rm B\kern-.05em{\sc i\kern-.025em b}\kern-.08em T\kern-.1667em\lower.7ex\hbox{E}\kern-.125emX}}

\newtheorem{definition}{Definition}
\newtheorem{lemma}{Lemma}
\newtheorem{theorem}{Theorem}

\title{Reductive Clustering: An Efficient Linear-time Graph-based Divisive Cluster Analysis Approach}
\author{
\IEEEauthorblockN{Ching Tarn\IEEEauthorrefmark{1}}
\and
\IEEEauthorblockN{Yinan Zhang\IEEEauthorrefmark{2}}
\and
\IEEEauthorblockN{Ye Feng\IEEEauthorrefmark{3}}
\thanks{\IEEEauthorrefmark{1}Key Laboratory of Intelligent Information Processing of Chinese Academy of Sciences, Institute of Computing Technology, Chinese Academy of Sciences, and University of Chinese Academy of Sciences. Email: i@ctarn.io.}
\thanks{\IEEEauthorrefmark{2}School of Computer Science and Engineering, Nanyang Technological University.}
\thanks{\IEEEauthorrefmark{3}School of Information Sciences, University of Illinois Urbana-Champaign.}
\thanks{The work was partly done while the authors were with Shandong University.}
}

\begin{document}
\maketitle

\begin{abstract}
We propose an efficient linear-time graph-based divisive cluster analysis approach called \textit{Reductive Clustering}.
The approach tries to reveal the hierarchical structural information through reducing the graph into a more concise one repeatedly.
With the reductions, the original graph can be divided into subgraphs recursively, and a lite informative dendrogram is constructed based on the divisions.
The reduction consists of three steps: selection, connection, and partition.
First a subset of vertices of the graph are selected as representatives to build a concise graph.
The representatives are re-connected to maintain a consistent structure with the previous graph.
If possible, the concise graph is divided into subgraphs, and each subgraph is further reduced recursively until the termination condition is met.
We discuss the approach, along with several selection and connection methods, in detail both theoretically and experimentally in this paper.
Our implementations run in linear time and achieve outstanding performance on various types of datasets.
Experimental results show that they outperform state-of-the-art clustering algorithms with significantly less computing resource requirements.
\end{abstract}

\begin{IEEEkeywords}
cluster analysis, hierarchical clustering, graph reduction, linear complexity
\end{IEEEkeywords}

\section{Introduction}
Cluster analysis is a basic common unsupervised learning approach and widely used in various fields such as data mining and computer vision.
It categorizes a set of objects into groups (or clusters) based on the similarities between them to help understand and analyze data.
Clusters are aggregations of objects that intra-cluster similarities are higher than inter-cluster ones.

With the long history of research on clustering, a large number of algorithms have been proposed.
However, there still exist several common limitations.

\subsection{Classical Partitional Methods}\label{sec:intro:partition}
Classical center-based algorithms like $k$-Means \cite{macqueen1967kmeans} use central vectors to partition the data space so that they may lack the ability to find arbitrary shaped clusters.
Many of them are also very sensitive to center initialization.
Distribution-based algorithms including Gaussian Mixture Models have similar shortcomings.
In addition to that, for spatial data, as the dimension grows, the number of parameters to be determined also increases.
The running time and memory required may become unacceptable.

Density-based clustering methods define clusters as contiguous regions with high density.
DBSCAN \cite{ester1996dbscan} assumes that the clusters have homogeneous densities.
A global threshold needs to be specified in advance to distinguish between high-density and low-density areas.
It simply judges whether two regions are connected based on connectivity, which can result in ``single-link effect" \cite{kriegel2011dbc} and lead to undesirable combinations.
Another critical problem is that, for some applications, there may not be a global threshold at all to separate all clusters at once.
It makes such tasks unsolvable to DBSCAN.

\begin{figure}
\centerline{\includegraphics[width=0.4\textwidth]{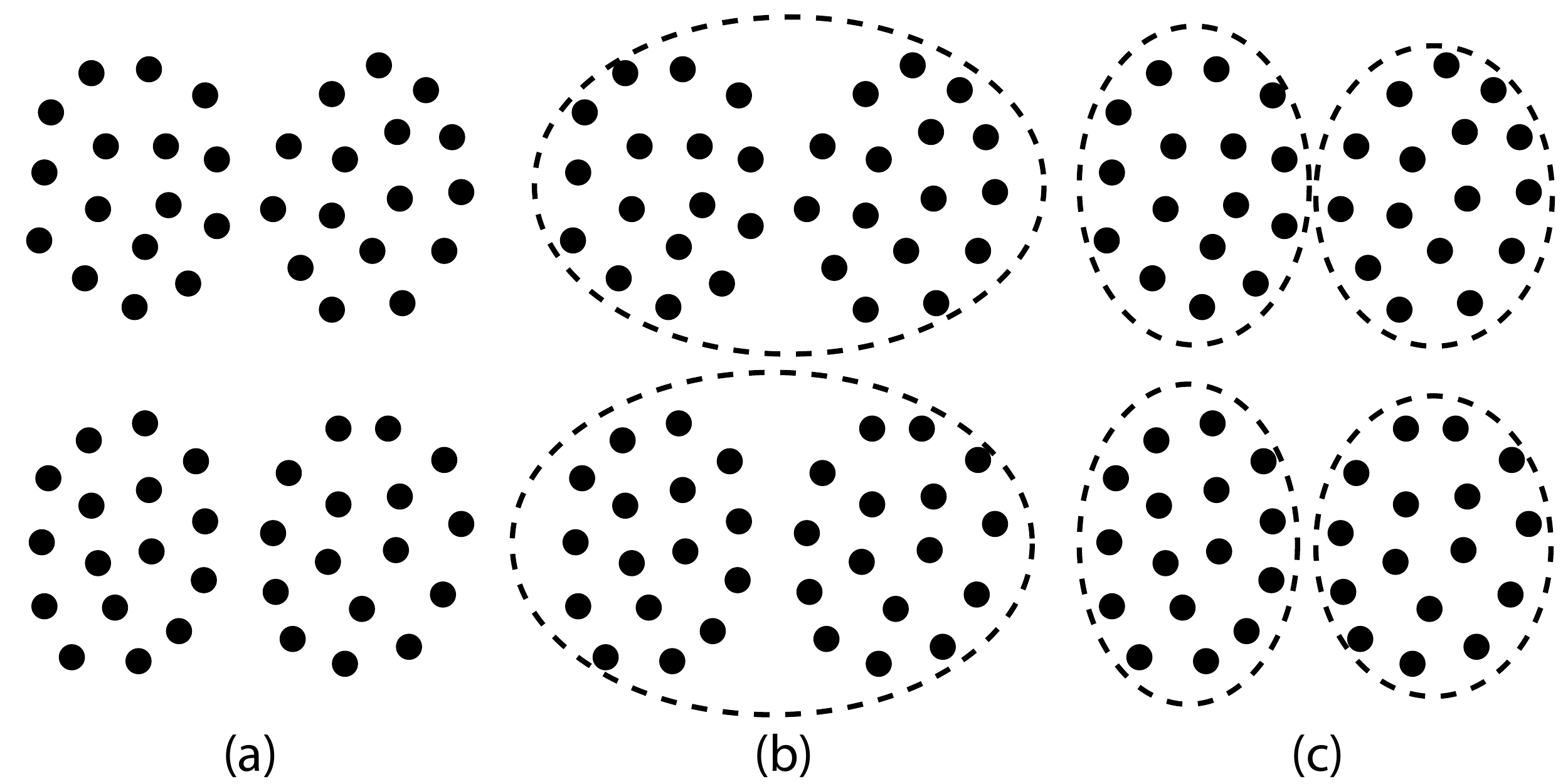}}
\caption{The figure indicates two partitions on a dataset.
    Obviously, since the data distribution is hierarchical, neither partition can fully represent the structure of the data.
    In order to obtain sufficient structural information, a partitional algorithm must be run many times.
    Moreover, it is also difficult to build a hierarchy from multiple partitions.
}
\label{fig:partition}
\end{figure}

Algorithms above generate disjoint partitions of the dataset, and nesting is not allowed.
However, in many cases, some data in reality are hierarchical.
As shown in Fig. \ref{fig:partition}, it is hard, or even impossible, to find an absolutely reasonable and unambiguous division of the data.
In such cases, a possible way is to generate multiple partitions on the same dataset with different granularities to construct a hierarchy.
However, it requires multiple runs and is difficult to handle.

\subsection{Agglomerative and Divisive Methods}
Hierarchical algorithms are traditionally divided into two categories, agglomerative methods and divisive ones.
Such algorithms generate hierarchical structures called dendrograms, which are much more informative than non-nested partitions.
When necessary, a dendrogram can also be converted into partitions with different numbers of clusters as needed without multiple runs.

Typically, agglomerative methods initially treat each object as a cluster, or start with a large number of tiny clusters, and then merge them in pairs.
The dendrogram is usually very deep and large, which makes it difficult to analyze.
In fact, in most cases, low level branches are not so meaningful, and often pruned to make the dendrogram easier to analyze, which is also a challenging task.

\begin{figure}
\centerline{\includegraphics[width=0.4\textwidth]{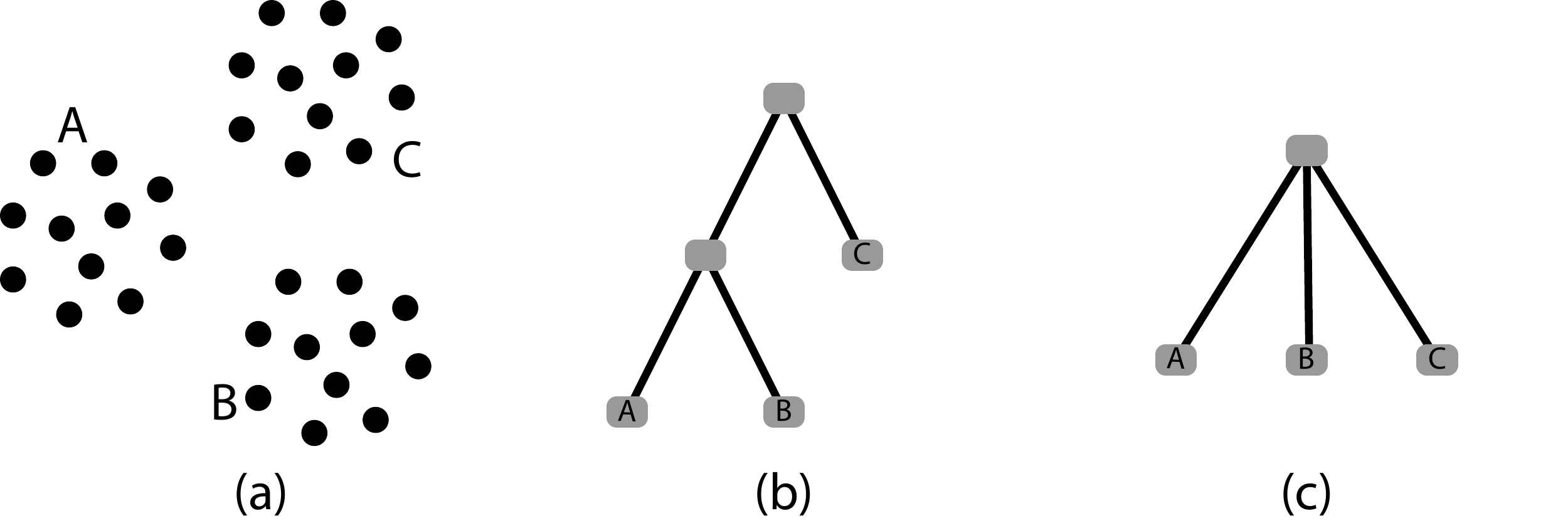}}
\caption{As can be seen, the dataset (a) consists of three similar parts: $A$, $B$ and $C$.
    Although the similarities between them are approximately equal, a binary dendrogram (b) may first groups $A$ and $B$ into one bigger cluster, which means that $A$ and $B$ are much more similar than $A$ and $C$ or $B$ and $C$.
    The non-binary dendrogram (c) is more desirable.
}
\label{fig:binary}
\end{figure}

Other than that, due to clusters being merged in pairs, dendrograms are usually binary and thus can not represent the real structures accurately.
For example, given a cluster consisting of more than two identical subclusters, there is no binary dendrogram that can correctly indicate the relationship between subclusters, as shown in Fig. \ref{fig:binary}.

For many agglomerative methods, to select a pair of clusters for the next merge, the similarities between every two clusters must be calculated.
The complexity of an agglomerative method is often very high, usually $O(N^3)$, which can be unacceptable for a mass of data.

In general, a divisive method initially groups all objects into one cluster, and divides each cluster into two subclusters recursively.

Compared to a large number of agglomerative algorithms, divisive methods are fewer.
One major reason is that separation is usually much more difficult than mergence.
For an agglomerative method, given $n$ subclusters, in order to find the two most similar ones to merge, only a maximum of $O(n^2)$ comparisons are required.
Correspondingly, there are $O(2^m)$ divisions that split a cluster of size $m$ into two subclusters.

Divisive methods have similar problems with agglomerative ones.
Besides that, the stopping condition (e.g., the commonly used \textit{minimum cluster size}) is also hard to be defined properly.
However, with the well-defined termination, a large number of unnecessary divisions can be avoided.
Unlike agglomerative methods, this feature makes it possible to generate a smaller and more analyzable dendrogram directly without any pruning.

\subsection{Graph-based Methods}\label{sec:intro:graph}
Many methods require the data to be spatial, which limits their applications.
Graph-based methods use similarity graphs, and usually transform spatial data into nearest neighbor graphs (e.g., $k$-NN, mutual $k$-NN \cite{brito1997mknn} and XNN \cite{franti2016xnn}).
Although the transformation may cause loss of information, processing on graphs still brings many significant benefits.
1) There are metrics in graph theory, like reachability, that can measure the similarities between objects better than traditional ones (e.g., the Euclidean distance).
2) Querying the most similar objects in a graph is often faster than others.
3) It has been shown in \cite{belkin2003laplacian} that nearest neighbor graphs are proper and efficient representations for high dimensional data lying on a low dimensional manifold.

\subsection{The Advantages of Reductive Clustering}
With the considerations introduced above and inspired by \cite{bar2016bpc}, we propose a graph-based divisive clustering approach called \textit{Reductive Clustering}.
It tries to reveal the structure of data at different granularities through repeatedly reducing the graph into a concise one.

The approach has following advantages.
\begin{itemize}
\item A non-binary dendrogram is generated, which is much more representative than non-nested partitions and binary dendrograms.
\item The division stops in time, and undesirable splits are less.
The dendrogram is small and easy to analyze.
\item It works well and outperforms state-of-the-art algorithms on various types of datasets.
\item Only a few hyper-parameters are required and none of them plays a key role.
\item Computing resources required are significantly less, and the time complexity is usually linear.
It is also easily parallelizable.
\item The approach is very simple, and can be extended to handle domain-specific data.
\end{itemize}

The paper is organized as follows.
We present details of the approach in Section \ref{sec:alg}, and compare it with other methods theoretically in Section \ref{sec:cmp}.
Section \ref{sec:exp} shows the experimental results.
\footnote{
    The related resources are at \url{http://res.ctarn.io/reductive-clustering}.
}
We discuss the experiments further in Section \ref{sec:discuss}.
The conclusions are in Section \ref{sec:con}.

\section{The Approach}\label{sec:alg}
\begin{figure}
\centerline{\includegraphics[width=0.4\textwidth]{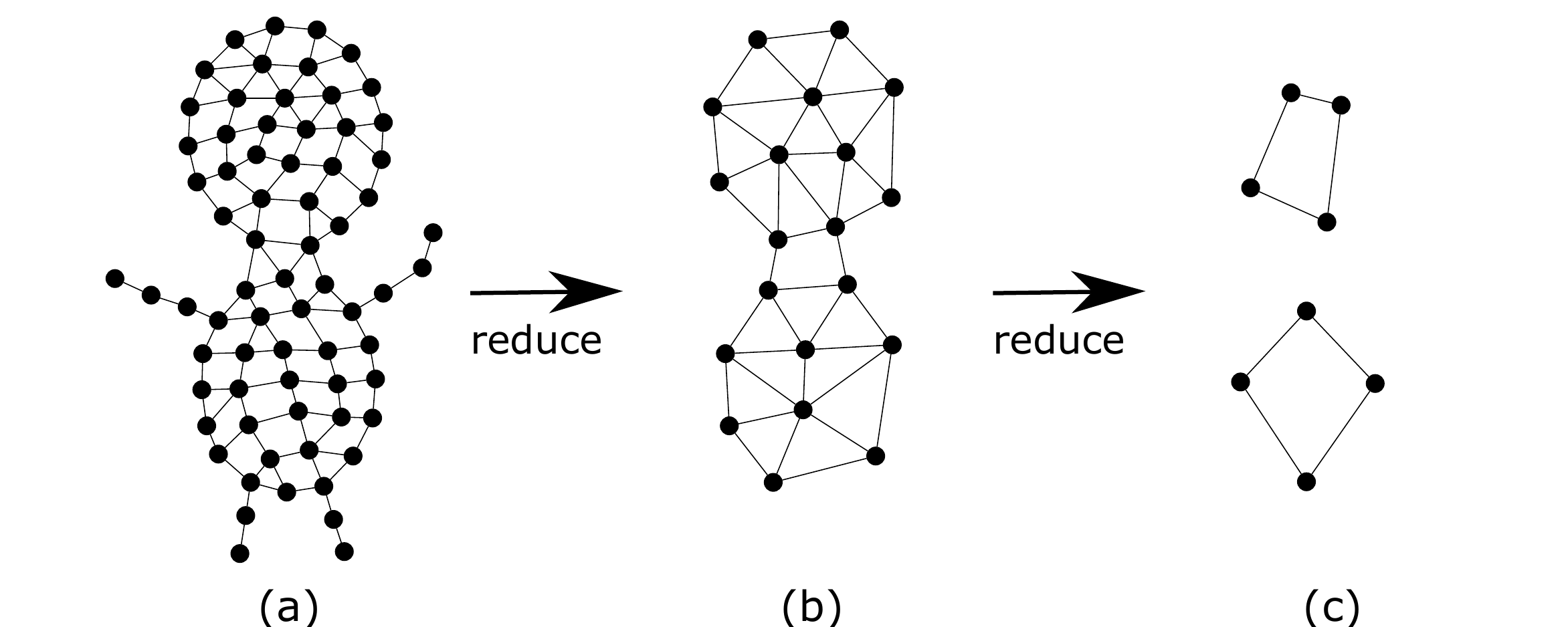}}
\caption{The figure shows that, after the first reduction, some less important details are removed from the graph, and the main body becomes more simple.
    Then it is further reduced, and the second reduction removes weak connections inside the graph, and clearly shows that the graph is made up of two parts.
}
\label{fig:reduction}
\end{figure}

\begin{figure*}
\centerline{\includegraphics[width=\textwidth]{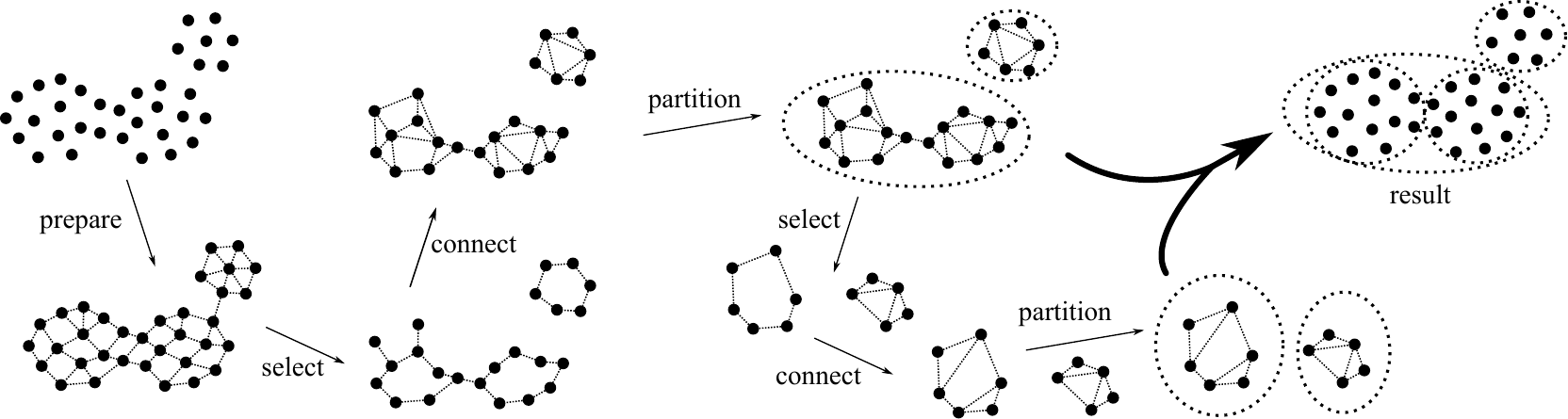}}
\caption{An ideal flow diagram.
    First the dataset is converted into a graph.
    Given a spatial dataset, the nearest neighbor graph can be used.
    The graph is reduced into a more concise graph.
    The reduction is preformed on each subgraph recursively until the termination condition is met.
    For selection, a subset of vertices are selected as representatives of the graph, and the others are removed.
    Each removed vertex is bound to a representative later, and the representative determines which cluster it belongs to.
    When the termination condition is met, the final representatives are grouped into clusters directly.
    After selection, representatives are reconnected tightly to avoid the fragmentation of a cluster and maintain a consistent structure of the graph.
    The reduced graph is divided into several subgraphs based on the connectivity.
}
\label{fig:flowchart}
\end{figure*}

The main idea of this approach is that, a big graph can be reduced into a more concise one that still retains the main structure but has fewer details.
The concise graph is easier to analyze, and can be further reduced again.
By reducing repeatedly, as shown in Fig. \ref{fig:reduction}, the structural information of different granularities can be gradually revealed.

With reduction, a connected graph can be divided into several disconnected subgraphs.
Like traditional methods, connectivity can be used as the criterion to group objects into clusters.
Each subgraph, that is, a connected component, is a subcluster of the previous cluster.
And thus, a hierarchy can be obtained by dividing recursively.

Reduction can be further divided into three steps: selecting a set of vertices from the original graph as the representatives, building a new graph by reconnecting these vertices, and dividing the concise graph into subgraphs.
A dendrogram can be built through dividing a graph into subgraphs recursively, as shown in Fig. \ref{fig:flowchart}.

Selection is applied to the graph first.
A subset vertices are selected as representatives, and unselected vertices are removed to disconnect latent subclusters.
Each unselected vertex is associated with a representative, and belongs to the same cluster with it.
The recursive reduction terminates if no vertex or all vertices of a graph are selected as representatives.

Connection is required to keep the structure of a graph stable.
Since the unselected vertices are removed from the graph, the connection between representatives are weak, and the remaining edges are usually not enough to connect them tightly to maintain a consistent structure.
With a large number of removals, it may even lead to the fragmentation of a graph.
By reconnecting, each representative is connected with several nearest representatives, and the structure can be well kept.

The last step of reduction is partition.
With vertices removals and edges adjustment, the connection between latent subclusters can be removed and the connected graph is divided into several separated components.
And thus a reduced graph can be partitioned based on the connectivity.

Moreover, for association, since each graph to be reduced is connected and there exists at least representative, it is guaranteed that each unselected vertex can always be associated to the nearest representative.
The initial graph is divided into subgraphs directly before reduction if it is not connected.

An overview is as Algorithm \ref{alg:rc}.

\begin{algorithm}
\caption{Reductive Clustering}
\label{alg:rc}
\begin{algorithmic}
\Procedure{Cluster}{$G = (V, E)$}
    \State $R \leftarrow \Call{Select}{G}$ \Comment representatives
    \If {$R = \emptyset$ \textbf{or} $R = V$}
        \State $U \leftarrow \text{the vertices associated with } V$
        \State \Return $U \cup V$ \Comment a leaf of the dendrogram
    \Else
        \ForAll {$v \in V \backslash R$}
            \State $r \leftarrow \text{the nearest vertex} \in R \text{ to } v$
            \State associate $v$ to $r$
        \EndFor
        \State $H \leftarrow \Call{Connect}{G, R}$ \Comment the new concise graph
        \State $D \leftarrow \emptyset$ \Comment an empty dendrogram node
        \ForAll {subgraph $S$ of $H$} \Comment partition
            \State $d \leftarrow \Call{Cluster}{S}$
            \State $D \leftarrow D \cup \{d\}$ \Comment append $d$ to $D$ as a child
        \EndFor
        \State \Return $D$
    \EndIf
\EndProcedure
\end{algorithmic}
\end{algorithm}

In the rest of this section, we describe the approach in more detail.
For reduction, we present a set of selection and connection methods, and introduce some measures to evaluate them to help choose or design one method that suits a specific application.
We also provide a fine-tuning method to further improve the quality of a dendrogram, and simple algorithms to convert a dendrogram into partitions.
In the end we briefly discuss the complexity of the approach.

Before that, given a directed graph $G = (V, E)$, a set of terms are defined as below.

\begin{definition}[neighbor]
    $u$ is a neighbor of $v$ if and only if $<v, u> \in E$.
\end{definition}

\begin{definition}[neighborhood]
    The neighborhood of $v$ is the set of all neighbors of $v$, denoted as $N(v)$.
    $$N(v) = \{u | <v, u> \in E\}$$
\end{definition}
In addition, $N(v) \cup \{v\}$ is denoted as $N^+(v)$.

\begin{definition}[dendrogram]
    A dendrogram of $G$ is a tree of which all leaves are exclusive subsets of $V$.
\end{definition}

\subsection{Graph Construction}\label{sec:alg:graph}
As described in Section \ref{sec:intro:graph}, transforming datasets into graphs brings many benefits.
In addition, graph is a flexible data structure, and can be easily edited or built.
Converting data in other forms into a graph is usually much easier than the reverse.
For example, building a spatial dataset based on similarity graphs is usually more difficult than transforming a spatial dataset into a similarity graph (e.g., $k$ nearest neighbor graph).
It enables a graph-based method to handle more types of data and thus applicable to more fields.

The graph construction methods depend on the original data type and the similarity metric used.
For general spatial data, in addition to brute-force search, fast indexing structures such as KD Tree \cite{bentley1975kdtree} or Ball Tree \cite{omohundro1989balltree} can be used to speed up the search of nearest neighbors.

Although the following discussions are mainly based on unweighted directed $k$ nearest neighbor graphs, the approach can also be applied to various type of graphs such as undirected or weighted graphs.
The methods introduced below may need to be slightly adjusted.

\subsection{Selection}\label{sec:alg:select}
Selecting a subset of vertices from a graph as representatives and removing unselected ones leads to several benefits.
1) By reducing the number of vertices, both analysis and computing become easier.
With a proper subset of the graph, the main structural information still can be well maintained.
2) Vertex removals can break the weak connections caused by noise between components.
It avoids common problems, like ``single-link effect", of linkage-methods and makes the approach more tolerant to noise.
3) More importantly, the separations of components reflect structural information of the data and can be used to reconstruct the hierarchical structure.

The selection algorithm is described as Algorithm \ref{alg:select}.
It uses a measure $s(\cdot)$ to calculate a score of each vertex that indicates whether it should be selected as a representative, and returns the set of vertices with the highest score, that is, the representatives.
We use selection rate $r$ to control the number of representatives.

\begin{algorithm}
\caption{Selection}
\label{alg:select}
\begin{algorithmic}
\Procedure{Select}{$G = (V, E)$}
    \ForAll {vertex $v \in V$}
        \State $v.score \leftarrow s(v)$
    \EndFor
    \State $n \leftarrow r |V|$ \Comment the number of representatives
    \State $t \leftarrow \text{$n$-th highest score}$ \Comment threshold
    \State $R \leftarrow \{v \in V | v.score \geq t \}$
    \State \Return $R$
\EndProcedure
\end{algorithmic}
\end{algorithm}

\begin{figure}
\centerline{\includegraphics[width=0.35\textwidth]{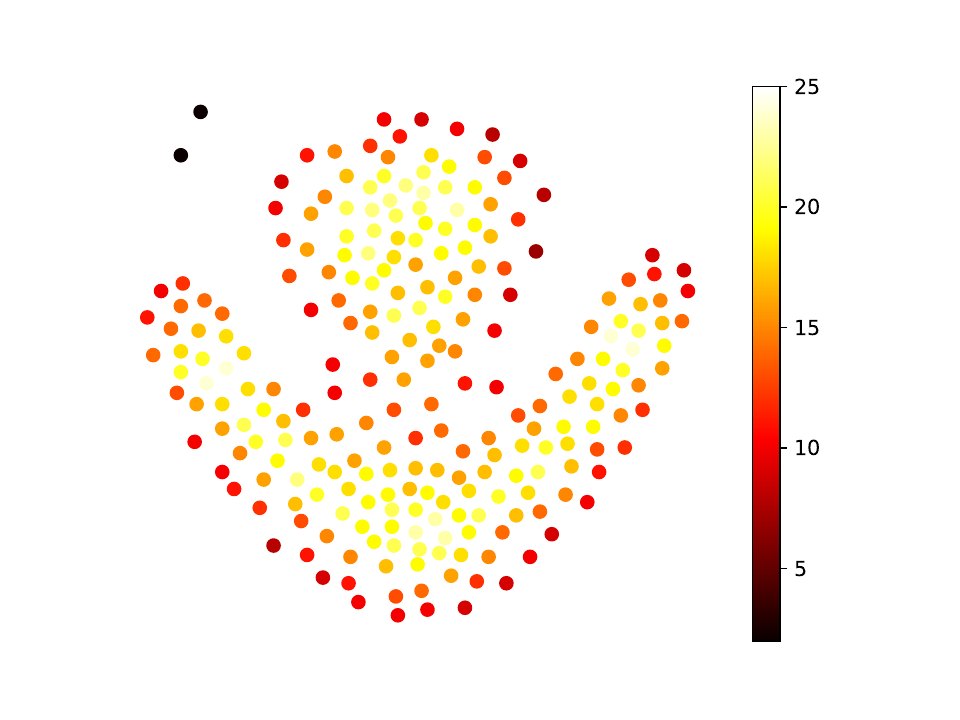}}
\caption{Indegree of each vertex on a 16-nn graph \cite{fu2007flame}.
    It shows that the indegree of a vertex on boundaries usually is smaller than others.
}
\label{fig:indegree}
\end{figure}

A basic selection measure $s_r$ is assigning a random score to each vertex, that is, selecting randomly.
Another type of methods tend to remove vertices at junctions of clusters instead of vertices in core areas, so that the latent clusters can be disconnected quickly and undesirable divisions can be avoided.
Such methods can be boundary detection based, like \cite{xiong2006border}.
It is shown in \cite{xiong2006border} that a vertex at junctions usually has a smaller indegree than others.
In particular, for a $k$-nn graph, the outdegree of each vertex is $k$, and the indegree of a vertex at boundaries is usually less than $k$, as shown in Fig. \ref{fig:indegree}.

\begin{definition}[indegree selection]\label{def:si}
    $$s_i(v) = |\{ u | v \in N(u) \}|$$
\end{definition}

Another similar one, defined as Definition \ref{def:sm}, identifies boundaries based on the number of mutual neighbors of each vertex.
\begin{definition}[mutual neighbor selection]\label{def:sm}
    $$s_m(v) = |\{ u | v \in N(u) \wedge u \in N(v) \}|$$
\end{definition}

To measure the effectiveness of a selection method, first we formally define the vertices at junctions as Definition \ref{def:pv}.
\begin{definition}[positive vertex]\label{def:pv}
    A vertex $v$ is at junctions if and only if $\exists u \in N(v)$ that $v$ and $u$ belong to different clusters.
    A vertex not at junctions is called a positive vertex.
\end{definition}

The reduced graph should be more separable.
We use the proportion of positive vertices of the graph to measure its separability.
\begin{definition}[vertex positivity]\label{def:Pv}
    The proportion of positive vertices of a graph is called its vertex positivity, denoted as $P_v$.
\end{definition}

\begin{figure}
\centerline{\includegraphics[width=0.5\textwidth]{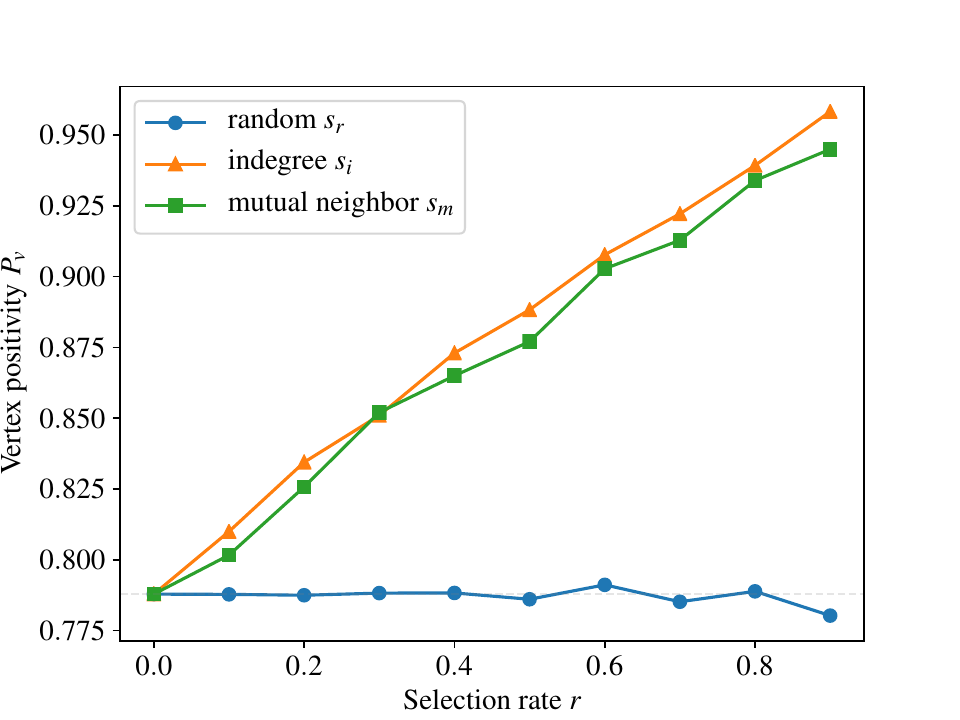}}
\caption{Vertex positivity evaluated with a 16-nn graph built from MNIST \cite{lecun1998cnn}.
    It shows that both $s_i$ and $s_m$ increase the vertex positivity of the graph significantly.
}
\label{fig:Pv}
\end{figure}

To avoid being affected by neighborhood size deceasing,
we label the positivity of a vertex before selection and only recalculate the proportion after that.
Obviously, selecting randomly doesn't change the proportion of positive vertices.
Fig. \ref{fig:Pv} shows that both $s_i$ and $s_m$ increase the vertex positivity of a graph significantly.

\subsection{Association}\label{sec:alg:associate}
\begin{figure}
\centerline{\includegraphics[width=0.45\textwidth]{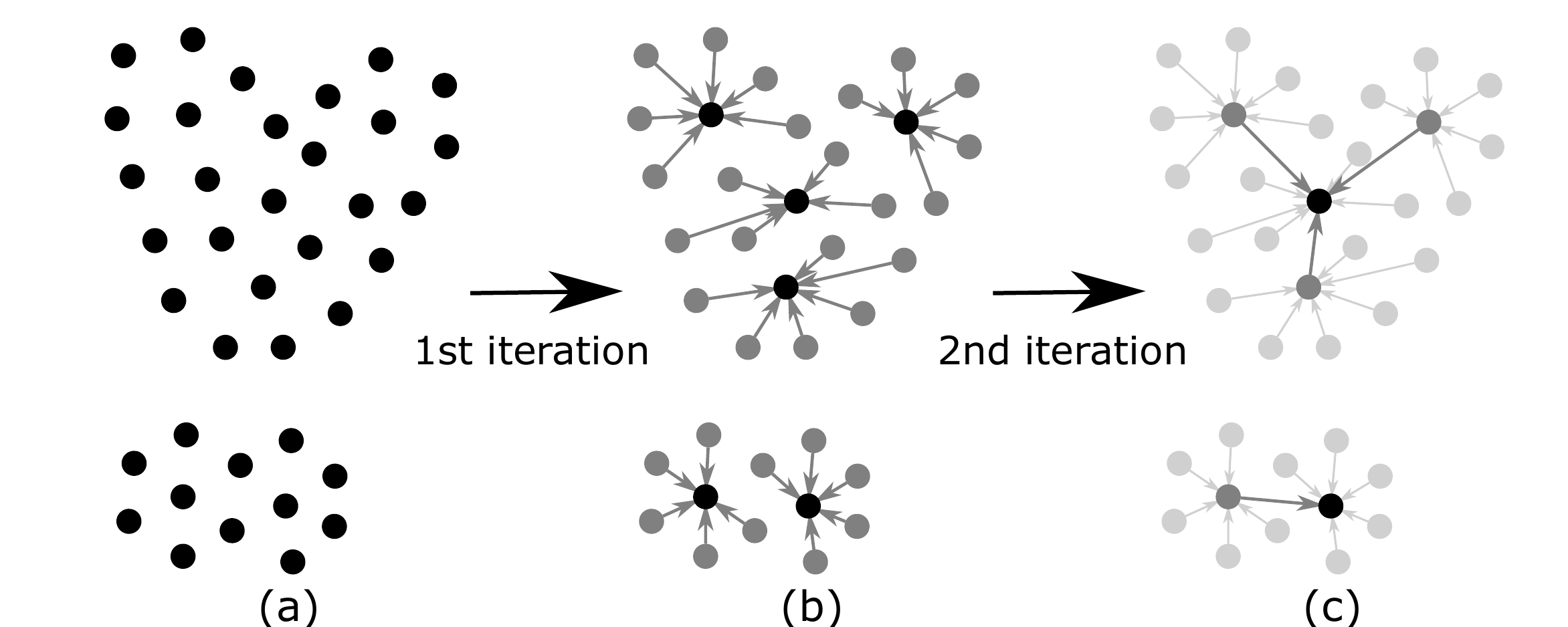}}
\caption{The figure shows that, the first selection removes all but 6 vertices, marked in gray.
    Each unselected one is associated with the nearest representative.
    After the second iteration, only 2 of the 6 vertices are selected as representatives, and the rest are associated with them.
    Assuming that the final two representatives are divided into two clusters, other ones are also grouped into the two clusters respectively.
}
\label{fig:associate}
\end{figure}

Since only the final representatives are grouped into clusters directly, in order to completely categorize all vertices, an unselected vertex should be associated with a representative, as shown in Fig. \ref{fig:associate}.
For each unselected one, there is always a final representative (indirectly) bound with it and they are grouped into the same cluster.

A simple method to query a nearest representative, described as Algorithm \ref{alg:bfs}, is breadth-first search starting from each unselected vertex, and, in general, it is also very fast.

\begin{algorithm}
\caption{Simple BFS Association}
\label{alg:bfs}
\begin{algorithmic}
\ForAll {vertex $v \in V \backslash R$} \Comment unselected vertices
    \State $Q \leftarrow \text{empty queue}$
    \State $\Call{Enqueue}{Q, v}$
    \While {$v$ has not been associated}
        \State $u \leftarrow \Call{Dequeue}{Q}$
        \ForAll {edges $<u, t>$}
            \If {$t \in R$} \Comment $t$ is a representative
                \State associate $v$ to $t$
                \State \textbf{break}
            \ElsIf {$t$ has not been visited}
                \State $\Call{Enqueue}{Q, t}$
            \EndIf
        \EndFor
    \EndWhile
\EndFor
\end{algorithmic}
\end{algorithm}

However, the worst-case performance of this method is $O(|V|^2)$.
For example, assuming a long chain of length $n$, and only the vertex at one end is marked as a representative, the number of visits on all edges is $\frac{n (n - 1)}{2}$.

In order to theoretically guarantee that the algorithm is linear in any case, we introduce another equivalent algorithm as an alternative.
Instead of using breadth-first search starting from each unselected vertex, we start breadth-first search from all representatives at the same time.
Since each edge is only visited at most once, it ensures that the search can be finished in linear time.
The details are shown as Algorithm \ref{alg:mbfs}.

\begin{algorithm}
\caption{Multi-source BFS Association}
\label{alg:mbfs}
\begin{algorithmic}
\State $Q \leftarrow \text{empty queue}$
\ForAll {vertex $v \in R$} \Comment representatives
    \State $v.label \leftarrow v$ \Comment label it as itself
    \State $\Call{Enqueue}{Q, v}$
\EndFor
\While {not all vertice have been labeled}
    \State $u \leftarrow \Call{Dequeue}{Q}$
    \ForAll {edges $<t, u>$} \Comment use reversed edges
        \If {$t$ has not been labeled}
            \State $t.label \leftarrow u.label$
            \State $\Call{Enqueue}{Q, t}$
        \EndIf
    \EndFor
\EndWhile
\ForAll {vertex $v \in V \backslash R$}
    \State associate $v$ to $v.label$
\EndFor
\end{algorithmic}
\end{algorithm}

\subsection{Connection}\label{sec:alg:connect}
A set of unselected vertices are removed from the graph after selection.
The graph can be much sparser and may even be broken into pieces.
We condense the graph by increasing the number of edges to avoid graph fragmentation caused by vertex removals.
Specifically, for a $t$ nearest neighbor graph, after selection and unselected ones have been removed, each representative is connected with $t$ newly found nearest vertices.

For a representative $v$, breadth-first search starting from $v$ is applied to find a set of candidates $C(v)$.
To avoid searching too deep unnecessarily, we limit the depth to $d$.
It stops if $|C(v)| \geq t$ and the searching depth is greater than $d$.
A measure $c(\cdot)$ is used to calculate a score of each candidate $u$ which indicates the similarity between $u$ and $v$.

The new neighborhood of a vertex consists of the candidates with the highest similarity.
A new graph is built through connecting each representative with their new neighbors.
The algorithm is described as Algorithm \ref{alg:connect}.

\begin{algorithm}
\caption{Connection}
\label{alg:connect}
\begin{algorithmic}
\Procedure{Connect}{$G = (V, E), R$}
    \State $E' \leftarrow \emptyset$ \Comment an empty edges set
    \ForAll {vertex $v \in R$}
        \State $C \leftarrow \emptyset$ \Comment candidates
        \State initialize a BFS iterator starting from $v$
        \While {$|C| < t$ \textbf{or} the depth of BFS $\leq d$}
            \State $u \leftarrow \text{the next vertex to be visited}$
            \If {$u = NULL$}
                \State \textbf{break} \Comment there is no reachable vertex.
            \ElsIf {$u \in R$}
                \State $u.score \leftarrow c(v, u)$
                \State $C \leftarrow C \cup \{ u \}$
            \EndIf
        \EndWhile
        \State $n \leftarrow \min(|C|, t)$ \Comment the actual number of neighbors
        \State $S \leftarrow$ the $n$ vertices in $C$ with the highest score
        \State $E' \leftarrow E' \cup \{ <v, u> | u \in S \}$ \Comment new edges
    \EndFor
    \State \Return $H = (R, E')$ \Comment new graph
\EndProcedure
\end{algorithmic}
\end{algorithm}

The simplest measure $c_g$ is based on geodesic distance, that is, the first $t$ vertices visited are selected as candidates.

Another measure is \textit{shared neighbor}, denoted as $c_s$.
It has been shown in \cite{ertoz2003sn} that a high similarity between two neighborhoods also indicates that the two vertices are similar.
Jaccard index, also known as Intersection over Union (IoU), can be used to measure the similarity between two neighborhoods.

\begin{definition}[Jaccard index]
    $$c_s(a, b) = \dfrac{|N^+(a) \cap N^+(b)|}{|N^+(a) \cup N^+(b)|}$$
\end{definition}

For spatial data, geometric distance metrics (e.g., Euclidean distance $c_d$) are also applicable in some applications.

Effective connection methods should maintain the structure of the graph well, and it can be measured from two aspects.
1) Inter-cluster connections should be avoided.
For the purpose of making clusters separable, a vertex should belong to the same cluster with its neighbors.
2) To avoid a cluster being divided into pieces, vertices in it need to be strongly connected.

For the first one, we use a similar measure with Definition \ref{def:Pv} called the edge positivity of a graph.

\begin{definition}[positive edge]\label{def:pe}
    An edge $<u, v>$ is positive if and only if $u$ and $v$ belong to the same cluster.
\end{definition}

\begin{definition}[edge positivity]\label{def:Pe}
    The proportion of positive edges of a graph is called its edge positivity, denoted as $P_e$.
\end{definition}

For the latter, the average connectivity of a graph \cite{beineke2002connectivity}, defined as follows, is used to measure the strength of a cluster's internal connections.
It can also be used to evaluate the stability of a selection method.

\begin{definition}[graph connectivity]\label{def:C}
    The connectivity of a graph is measured with
    $$C = \frac{\sum\limits_{u, v \in V} f(u, v)}{{|V| \choose 2}},$$
    where $f(u, v)$ is the value of the maximum flow from $u$ to $v$.
\end{definition}

\begin{figure}
\centerline{\includegraphics[width=0.5\textwidth]{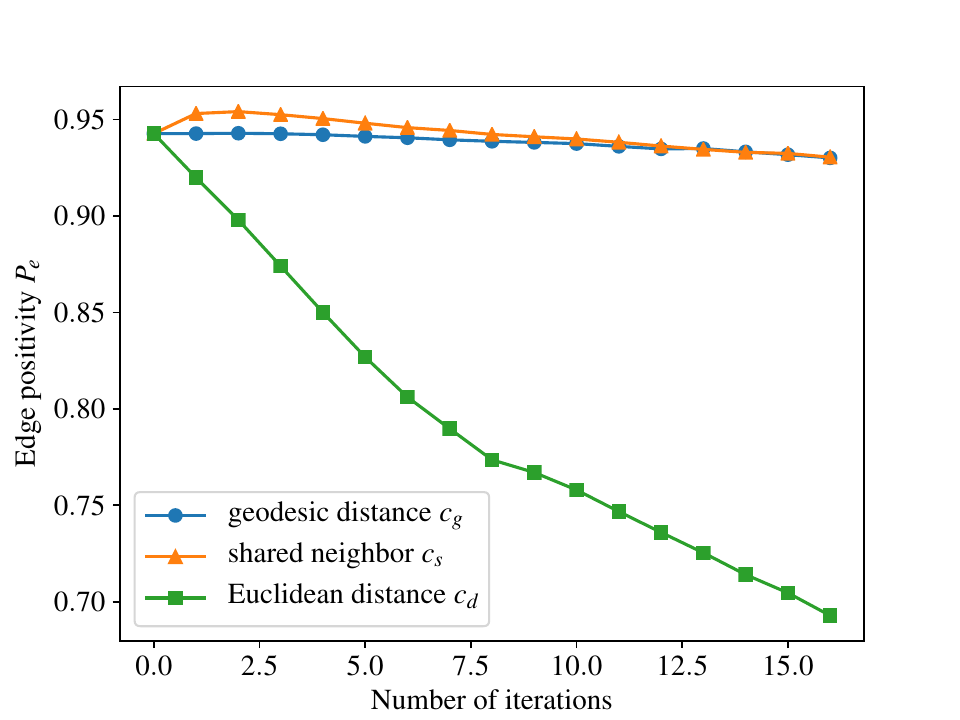}}
\caption{Edge positivity evaluated with a 16-nn graph built from MNIST.
    The connection methods based on different measures are applied on the graph independently with $t = 16$.
    In each iteration, $10 \%$ vertices are randomly removed from the graph.
    It shows that both $c_g$ and $c_s$ are significantly more effective than $c_d$.
}
\label{fig:Pe}
\end{figure}

Fig. \ref{fig:Pe} shows that graph-based measures are usually more effective than geometry-based.

Although $c_s$ is very effective, it is not a stable connection method on a big graph.
With the reconnection being repeated multiple times on a graph, it tends to divide a component into lots of tiny parts.
The size of such a part is usually about $t$, and the vertices of each part are almost fully-connected, while inter-part connections are extremely weak.
The intra-cluster connections are destroyed and the clusters are divided into pieces.

\begin{figure}
\centerline{\includegraphics[width=0.5\textwidth]{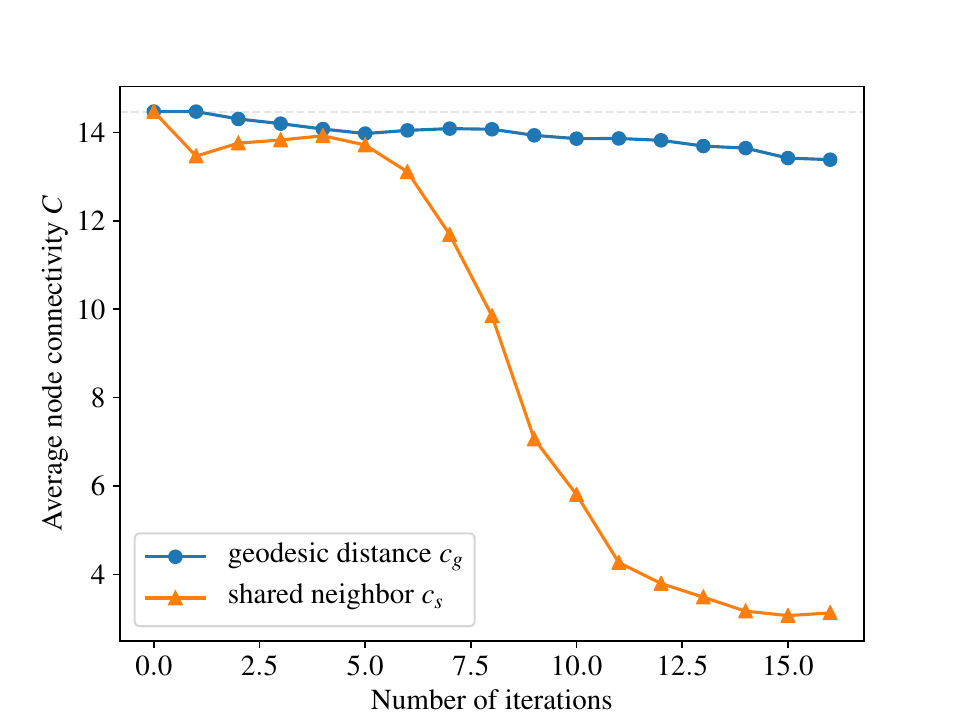}}
\caption{Connectivity evaluated with a random graph.
    The graph contains $256$ vertices, and each is randomly connected to $16$ other ones.
    In each iteration, connection is applied with $t = 16$, and $10 \%$ vertices are randomly removed from the graph.
    It shows that the connection method based on $c_g$ can maintain the connectivity well.
    For the $c_s$ based method, the connectivity decreases rapidly after about $6$ iterations.
}
\label{fig:C}
\end{figure}

Fig. \ref{fig:C} shows that, the connectivity of a graph decreases rapidly after about $6$ iterations if $c_s$ is employed.
In fact, if a $k$-nn graph is reduced without any vertex being removed, tiny parts are generated even faster, while $c_g$ doesn't change the graph at all if $t = k$.

\subsection{Partition}\label{sec:alg:partition}
At the end of each reduction, the new concise graph is partitioned into several subgroups, which generates a branch on the dendrogram.
Since the similarities between elements can be measured by reachability, the graph can be simply divided into connected components.
For data difficult to partition, in order to split it more thoroughly, the graph is divided into strongly-connected components.
Otherwise, or the graph is undirected, weakly-connected components methods are also applicable.

Additionally, dividing a graph into components also makes sure that each subgraph to be further reduced is connected, which guarantees that each unselected vertex can always be associated to a representative.

\subsection{Pruning}\label{sec:alg:prune}
Pruning is used to reduce the number of leaves of a dendrogram and generate a partition from it.
We provide two simple pruning algorithms.

The first algorithm strictly conforms to the original structure of a dendrogram.
A node can be merged if and only if it is an end branch, defined as Definition \ref{def:end-branch}.

\begin{definition}[end branch]\label{def:end-branch}
    An end branch of a dendrogram is a node of the dendrogram whose all children are leaves.
\end{definition}

The size of a node is defined as the total number of objects in its descendants (including itself).

The smallest end branch is first merged into one leaf.
The algorithm stops if the next merge causes the number of leaves to be less than the desired value $n$, or the largest $n$ leaves have contained more than $\alpha|V|$ objects, described as Algorithm \ref{alg:hard-pruning}.

\begin{algorithm}
\caption{Hard Pruning}
\label{alg:hard-pruning}
\begin{algorithmic}
\While {the sum of the sizes of largest $n$ leaves $< \alpha |V|$}
    \State $B \leftarrow \text{the smallest end branch}$
    \State $n_l \leftarrow \text{the number of leaves}$
    \State $n_b \leftarrow \text{the number of leaves of $B$}$
    \If {$n_l - n_b + 1 < n$}
        \State \textbf{break} \Comment the number of leaves is small enough
    \EndIf
    \State $L \leftarrow \emptyset$ \Comment new empty leaf
    \ForAll {leaf $l$ of $B$}
        \State $L \leftarrow L \cup l$ \Comment move objects to the new leaf
    \EndFor
    \State replace $B$ with $L$
\EndWhile
\end{algorithmic}
\end{algorithm}

\begin{figure}
\centerline{\includegraphics[width=0.45\textwidth, trim=0 70 0 0, clip]{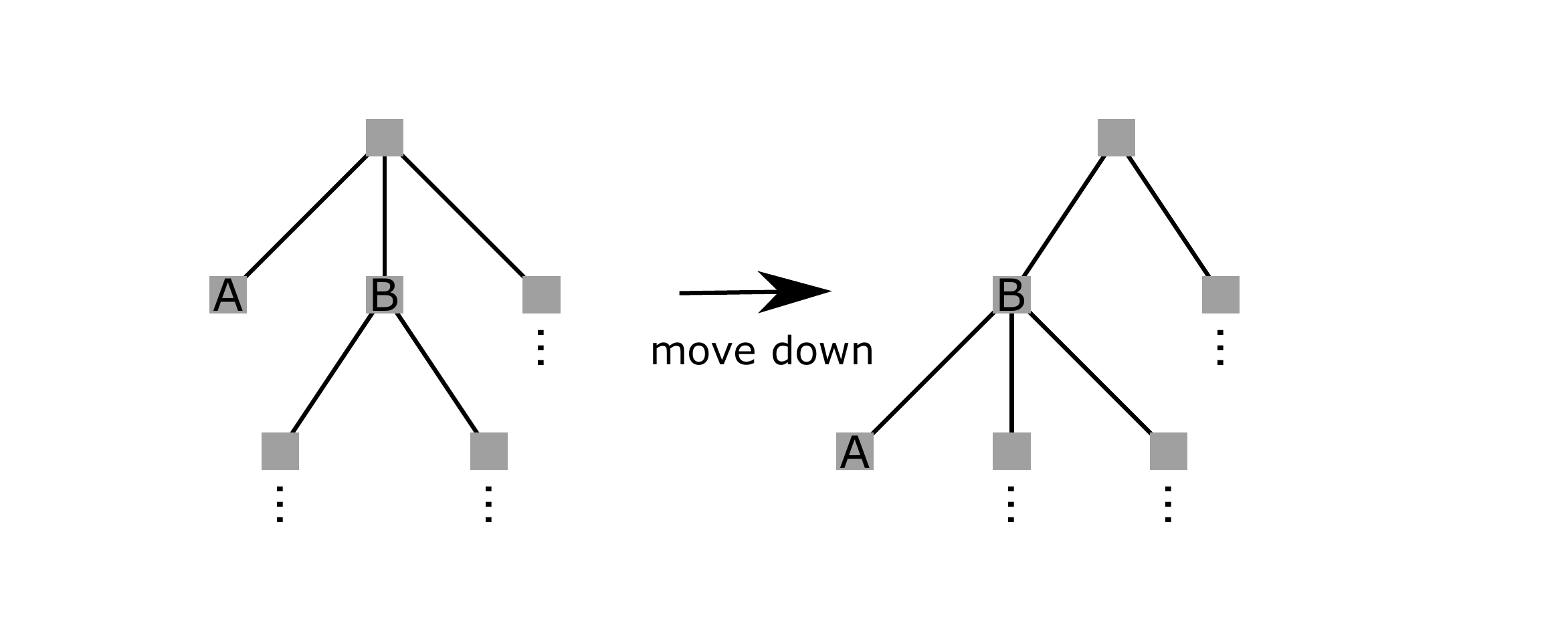}}
\caption{Assuming that $A$ is the smallest leaf of the dendrogram, and $B$ is the smallest brother of $A$.
    Since $B$ is a branch and can not be merged with $A$ directly, $A$ is moved into $B$ so that it may be able to merged with a leaf child of $B$.
    The leaf $A$ may be moved down multiple times, until its smallest brother is a leaf.
}
\label{fig:soft-prune}
\end{figure}

The second algorithm allows two leaves belonging to the same parent to be merged together directly.
A leaf is moved down to find a leaf brother, as shown in Fig. \ref{fig:soft-prune}, if its smallest brother is not a leaf.
The detailed description is shown in Algorithm \ref{alg:soft-pruning}.
It generates a more balanced partition with the number of clusters being precisely controlled.
However, it dose not strictly conform to the original structure of the dendrogram and doesn't work well on unbalanced data.

\begin{algorithm}
\caption{Soft Pruning}
\label{alg:soft-pruning}
\begin{algorithmic}
\While {$\text{the number of leaves} > n$}
    \State $l \leftarrow \text{the smallest leaf}$
    \State $p \leftarrow l.parent$
    \If {$p$ has more than one child}
        \State $b \leftarrow \text{the smallest brother of $l$}$
        \If {$b$ is a leaf}
            \State merge $l$ and $b$
        \Else \Comment $b$ is a branch
            \State move $l$ into $b$
        \EndIf
    \EndIf
    \If {$p$ has only one child}
        \State replace $p$ with its child
    \EndIf
\EndWhile
\State \Return $D$
\end{algorithmic}
\end{algorithm}

\subsection{Fine-tuning}\label{sec:alg:fine-tune}
We also introduce a simple method called \textit{smoothing} to adjust dendrograms.
It is based on the observation that, firstly, some association trees, which are generated after selection, may cross together slightly, which results in the fact that a small number of vertices, especially at junctions, are grouped into different clusters from their neighbors;
secondly, a few isolated vertices may become clusters unexpectedly, and such tiny leaves should also be removed.

We simply regroup each vertex into the most-common cluster in its neighborhood.
This operation can be repeated multiple times, and the experiments show that it can improve dendrograms on most datasets and also converges very fast.

\subsection{Complexity Analysis}\label{sec:complexity}
At the end, we briefly analyze the complexity of the approach.
Given a $k$-nn graph $G = (V, E)$, first we discuss the complexity of each step of reduction.

\subsubsection{Selection}
Assuming a selection measure $s(\cdot)$ with a complexity of $f_s$, calculating the scores costs $O(|V| f_s)$.
Finding the threshold can also be finished at the cost of $O(|V|)$.
Therefore, the complexity of selection is $O(|V| f_s)$.

\begin{lemma}
    The complexity of selection is $O(|V| f_s)$.
\end{lemma}

\subsubsection{Association}
As described in Section \ref{sec:alg:associate}, the worst-case performance of multi-source BFS association is $O(|V|)$.

\begin{lemma}
    The complexity of association is $O(|V|)$.
\end{lemma}

\subsubsection{Connection}
With the searching depth being limited to $d$, for each vertex, the number of candidates does not exceed $\max(t, \sum\limits_{i = 1}^{d}{k^i})$,
and thus both calculating and sorting the scores cost only $O(f_c)$, where $f_c$ is the complexity of the measure.

\begin{lemma}
    The complexity of connection is $O(|V| f_c)$.
\end{lemma}

\subsubsection{Partition}
Methods to computing strongly or weakly connected components also run in linear time.

\begin{lemma}
    The complexity of partition is $O(|V|)$.
\end{lemma}

Therefore, the complexity of reduction is $O(|V|(f_s + f_c))$.

Since the total size of all subgraphs is $r |V|$, where $r$ is the selection rate and $0 < r < 1$, the complexity of the approach is $O(|V| (f_s + f_c))$.

\begin{theorem}
    The complexity of Reductive Clustering is $O(|V| (f_s + f_c))$.
\end{theorem}

The complexity of any selection or connection measure introduced above is $O(1)$, and thus the complexity of an implementation based on them is linear.

\begin{theorem}
    In the case where both the costs of selection and connection measures are $O(1)$, the complexity of Reductive Clustering is $O(|V|)$.
\end{theorem}

\section{Related Work}\label{sec:cmp}
We compare \textit{Reductive Clustering} with other cluster analysis methods theoretically in this section.

Almost all existing algorithms, including the classical methods mentioned above, can be classified into two categories.
For the first type, the methods process on static original data distributions or graphs.
For the second one, the methods are usually iterative, and dynamically adjust the distribution of all objects or the structure of a graph.

\subsection{Methods Processing on Static Distributions}
Global characteristics matter.
Methods using static distributions often lack the ability to reveal global characteristics.
A typical example is DBSCAN \cite{ester1996dbscan}.
As mentioned in Section \ref{sec:intro:partition}, since the method simply groups objects into a cluster based on local connectivity, DBSCAN often merges independent clusters by mistake.
Other derivative algorithms of DBSCAN, like HDBSCAN \cite{campello2013hdbscan}, do not solve it very well either.
Iterative algorithms based on static modeling, including $k$-Means \cite{macqueen1967kmeans}, $k$-Means++ \cite{arthur2007kmeans++}, $k$-Medoid \cite{kaufman1987kmedoids}, $k$-Medians \cite{jain1988kmedians}, Gaussian Mixture Models and BRICH \cite{zhang1996birch}, perform better in this regard.
However, since the modeling abilities are also limited, most of them are only suitable for specific types of distribution, which are mainly convex structures.
Additionally, many of them, especially for BRICH, are sensitive to parameters, and thus require a good understanding of the data.

Highly Connected Subgraphs \cite{hartuv2000hcs} (HCS), is a positive example.
It recursively splits a graph into two subgraphs based on minimum cuts, which is a good global measure.
Although it doesn't work through either \cite{shi2000ncuts}, HCS is more tolerant to noise.
However, it is obvious that computing minimum cuts multiple times is a time consuming task.

Chameleon Clustering \cite{george1999chameleon} is an effective agglomerative method.
It groups data into a large number of tiny subclusters, and repeatedly merges two clusters that are relatively close and interconnected.
Unfortunately, the metrics used to measure the similarity and interconnectivity between pairs of clusters only perform well in low-dimensional spaces.

Graph Degree Link \cite{zhang2012gdl} (GDL), is another graph-based agglomerative algorithm.
It uses indegree and outdegree to measure the similarities between clusters, and merges them in pairs.
The parameters are usually difficult to be specified properly, and often require multiple runs to find usable settings.
However, in practice, it is often hard to evaluate the quality of a clustering result without ground-truth labels, which results in the fact that there is no reliable method to measure the parameter settings.
Moreover, just like most agglomerative methods, it is also very time consuming and not applicable on big datasets.

\subsection{Methods Adjusting Distributions Dynamically}
Robust Continuous Clustering \cite{shah2017rcc} (RCC), is an iterative method and expresses clustering as optimization of a continuous objective.
The method associates each data point with a representative, and optimizes them to reveal the structure of data distribution.
In the optimization process, the representatives gradually gather into several clusters.
It is fast and also works well in high-dimensional spaces.
Another notable feature is that the number of clusters need not to be specified in advance.
However, it also causes the granularity of clustering to be uncontrollable.
Even worse, since the optimized distribution can not be further easily interpreted, it is almost helpless to split or merge existing clusters manually.

There are also a set of cluster analysis methods based on dimensionality reduction.
They either require a lot of computing resources (e.g., t-SNE \cite{maaten2008tsne}), or perform poorly.

\subsection{Reductive Clustering}
As far as we know, the algorithm schema proposed in this paper, \textit{Reductive Clustering}, is the first approach that adjusts the structure of a graph dynamically through reducing the graph into a concise one repeatedly.
It makes the global characteristics to be well revealed easily with significantly less computing resources required.

\section{Experiments}\label{sec:exp}
\subsection{Comparison with Other Algorithms}\label{sec:exp:alg}
\subsubsection{Datasets}
We follow the experiments in \cite{shah2017rcc}, Robust Continuous Clustering, and use the datasets preprocessed and publicly provided by the authors.

For YaleB \cite{georghiades2001yaleb}, we only use the frontal face images processed using gamma correction and DoG filter.
For TYF \cite{wolf2011yft}, the video frames of the first 40 subjects sorted in chronological order are used.
For Reuters-21578, the train and test sets of the Modified Apte are used, and categories with less than five instances are not considered.
For RCV1 \cite{lewis2004rcv1}, the target clusters are defined as four root categories.
We use a random subset of 10,000 instances.
For text datasets, Reuters and RCV1, only the 2,000 most frequently occurring word stems are considered.
There is no additional preprocessing for other datasets.
Unlike \cite{shah2017rcc}, for methods other than Robust Continuous Clustering, the features are not normalized or reduced to a low dimension.

A brief summary is shown in Table \ref{tab:datasets}.

\begin{table}
\caption{Datasets}
\label{tab:datasets}
\begin{center}
\begin{tabular}{ l r r r }
\hline
\textbf{Dataset}
& \textbf{Instances} & \textbf{Dimensions} & \textbf{Classes} \\ \hline
\textbf{MNIST\cite{lecun1998cnn}}
& 70,000             & 784                 & 10  \\
\textbf{COIL100\cite{nene1996coil100}}
& 7,200              & 49,152              & 100 \\
\textbf{YaleB\cite{georghiades2001yaleb}}
& 2,414              & 32,256              & 38  \\
\textbf{YTF\cite{wolf2011yft}}
& 10,056             & 9,075               & 40  \\
\textbf{Reuters}
& 9,082              & 2,000               & 50  \\
\textbf{RCV1\cite{lewis2004rcv1}}
& 10,000             & 2,000               & 4   \\
\textbf{Pendigits\cite{alimoglu1997pendigits}}
& 10,992             & 16                  & 10  \\
\textbf{Shuttle}
& 58,000             & 9                   & 7   \\
\textbf{Mice Protein\cite{higuera2015miceprotein}}
& 1,077              & 77                  & 8   \\
\hline
\end{tabular}
\end{center}
\end{table}

\subsubsection{Baselines}
We compare Reductive Clustering (RC) with both partitional methods,
including $k$-Means++ (KM) \cite{arthur2007kmeans++}, Mean Shift (MS) \cite{comaniciu2002meanshift}, Gaussian Mixture Models (GMM), Affinity Propagation (AP) \cite{frey2007ap} and Robust Continuous Clustering (RCC, RCC-DR),
and hierarchical methods,
including three classical agglomerative algorithms (AC-Average, AC-Complete, AC-Ward), Graph Degree Linkage (GDL-U, AGDL) \cite{zhang2012gdl} and Hierarchical DBSCAN (HD) \cite{campello2013hdbscan}.

For RCCs (RCC, RCC-DR) and GDLs (GDL-U, AGDL), we use implementations publicly provided by the authors, and for others, we use scikit-learn and scikit-learn-contrib.

\subsubsection{Measures}
We use $\text{AMI}_\text{max}$ \cite{vinh2010measure} provided by scikit-learn to evaluate all algorithms.
A score varies in range $[0, 1]$, and the higher the better.

\subsubsection{Settings}\label{sec:exp:setting}
\begin{description}
\item[RC]
The Euclidean distance metric is used to build a $16$-nn graph of each dataset.
We use indegree selection $s_i$ and the geodesic distance based connection $c_g$.
Although $c_s$ is not stable and may divide a graph into pieces after multiple runs, it still be usable and very effective, and thus we run connection using $c_s$ first only once to improve the quality of the graph.
The parameters are fixed as $r = 0.8$, $t = 16$.
Additionally, the unselected vertices are removed from the graph before connection, and the search depth is limited to $2$.
Graphs are divided into strongly connected components.
Smoothing runs $16$ times before and after pruning.
We use hard pruning method with $\alpha$ being fixed as $0.8$ on unbalanced or noise-rich datasets (YTF, Reuters, RCV1, Shuttle and  Mice Protein) and soft pruning on others.
\item[KM]   Run each algorithm $4$ times.
\item[MS]   $\textit{quantile} \in \{0.001, 0.003, 0.01, 0.03, 0.1\}$.
\item[GMM]  Run each algorithm $4$ times.
\item[AP]   $\textit{max iter} = 1000$, $\textit{convergence iter} = 100$, $\textit{damping} = 0.9$.
\item[RCCs] $\textit{max iter} = 100$, $\textit{inner iter} = 4$. The weighted graphs are provided by the authors.
\item[GDLs] $a \in \{10^{-2}, 10^{-1.5}, \cdots, 10^{2}\}$, $K = 20$.
\item[HD]   $\textit{min cluster size} \in \{ 2^1, 2^2, \cdots, 2^{\lfloor log_{2}{\bar{n}} \rfloor} \}$, where $\bar{n}$ is the average size of the ground-truth clusters.
\end{description}

The default settings are used if not mentioned.
For algorithms that run multiple times, including KM, GMM, MS, GDLs and HD, the best results are reported.

\subsubsection{Results}
We use a computer with an Intel Core i7-6770HQ CPU ($2.60\text{GHz} \times 8$) and $31.3$ GiB memory, and running Ubuntu Desktop 18.04.
The results are shown in Table \ref{tab:AMI}.
Some algorithms may require too much memory on a dataset and thus are not applicable, marked as \textit{MLE}.

The computing resource costs are also compared.
Considering the scalability of some algorithms, as shown in Table \ref{tab:costs}, we only evaluate them on RCV1 to make the results complete.

\begin{table*}
\caption{Results measured by AMI}
\begin{center}
\begin{tabular}{l c c c c c c c c c c c c c c}
\textbf{Dataset}
& \textbf{KM} & \textbf{MS} & \textbf{GMM} & \textbf{AP} & \textbf{RCC} & \textbf{RCC-DR} &
& \textbf{AC-A} & \textbf{AC-C} & \textbf{AC-W} & \textbf{GDL-U} & \textbf{AGDL} & \textbf{HD} & \textbf{RC} \\ \hline
\textbf{MNIST}
& 0.496 & 0.226 & 0.281 &  MLE  & 0.869 & 0.746 &  &  MLE  &  MLE  &  MLE  &  MLE  &  MLE  & 0.189 & \textbf{0.895} \\
\textbf{COIL100}
& 0.793 & 0.706 &  MLE  & 0.635 & 0.924 & 0.924 &  & 0.514 & 0.650 & 0.825 & \textbf{0.936} & \textbf{0.936} & 0.860 & 0.862 \\
\textbf{YTF}
& 0.775 & 0.714 &  MLE  & 0.578 & 0.788 & 0.783 &  & 0.429 & 0.621 & \textbf{0.803} & 0.576 & 0.563 & 0.758 & 0.791 \\
\textbf{YaleB}
& 0.593 & 0.272 &  MLE  & 0.526 & \textbf{0.958} & \textbf{0.958} &  & 0.106 & 0.387 & 0.726 & 0.955 & 0.955 & 0.618 & 0.900 \\
\textbf{Reuters}
& 0.381 & 0.000 & 0.384 & 0.198 & 0.379 & 0.398 &  & \textbf{0.462} & 0.289 & 0.357 & 0.431 & 0.429 & 0.282 & 0.376 \\
\textbf{RCV1}
& 0.506 & 0.000 & \textbf{0.556} & 0.129 & 0.106 & 0.365 &  & 0.057 & 0.106 & 0.306 & 0.066 & 0.144 & 0.181 & 0.160 \\
\textbf{Pendigits}
& 0.665 & 0.680 & 0.712 & 0.427 & 0.730 & 0.800 &  & 0.566 & 0.557 & 0.707 & 0.422 & 0.422 & 0.699 & \textbf{0.857} \\
\textbf{Shuttle}
& 0.136 & 0.267 & 0.223 &  MLE  & 0.290 & 0.365 &  & 0.010 & 0.011 & 0.160 &  MLE  &  MLE  & \textbf{0.597} & 0.293 \\
\textbf{Mice Protein}
& 0.457 & 0.438 & 0.416 & 0.384 & 0.500 & 0.520 &  & 0.239 & 0.274 & 0.486 & 0.403 & 0.403 & 0.379 & \textbf{0.526} \\
\hline
\multicolumn{15}{l}{MLE: Memory Limit Exceeded.}
\end{tabular}
\end{center}
\label{tab:AMI}
\end{table*}

\begin{table*}
\caption{Computing resource costs on RCV1}
\begin{center}
\begin{tabular}{l c c c c c c c c c c c c c c}
\textbf{Costs on RCV1}
& \textbf{KM} & \textbf{MS} & \textbf{GMM} & \textbf{AP} & \textbf{RCC} & \textbf{RCC-DR} &
& \textbf{AC-A} & \textbf{AC-C} & \textbf{AC-W} & \textbf{GDL-U} & \textbf{AGDL} & \textbf{HD} & \textbf{RC} \\ \hline
\textbf{Time (sec)}
& 25 & 114 & 187 & 757 & 6906 & 458 & & 64 & 64 & 64 & 26 & 21 & 359 & 22 \\
\textbf{Memory (MiB)}
& 479 & 1279 & 1055 & 4017 & 1983 & 1356 & & 935 & 935 & 935 & 1807 & 1804 & 912 & 686 \\
\hline
\multicolumn{15}{l}{Time = Elapsed time $\times$ Percent of CPU. Memory = Maximum resident set size.} \\
\multicolumn{15}{l}{The costs of computing graphs (RCCs, GDLs and RC), distance matrices (GDLs) and bandwidths (MS) are not included.} \\
\multicolumn{15}{l}{The RCCs and GDLs are implemented in Matlab and others in Python. RC is mainly based on scipy, numpy and networkx.}
\end{tabular}
\end{center}
\label{tab:costs}
\end{table*}

It shows that the implementation based on \textit{Reductive Clustering} achieves the best results on three datasets and requires significantly less computing resources, while every other algorithms, including state-of-the-art algorithms like RCCs and GDLs, works best on at most one dataset.

GMM failed on high dimensional datasets.
MNIST is the biggest dataset, and all agglomerative methods and affinity propagation are not applicable on it.
GDLs and AP also failed on another big dataset, Shuttle.

We discuss the results further in Section \ref{sec:discuss}.

\subsection{Selection and Connection Measures}
We evaluate the measures on four datasets, including MNIST, Pendigits, YaleB and YTF.
The results are shown in Table \ref{tab:measures}.
The $\bar{n}$ indicates the average size of the ground-truth clusters of each dataset.

\begin{table}
\begin{center}
\caption{Comparison of measures}
\label{tab:measures}
\begin{tabular}{c c c c c c c c c}
\multicolumn{4}{c}{MNIST ($\bar{n} = 7,000$)} & &
\multicolumn{4}{c}{Pendigits ($\bar{n} \approx 1,099$)} \\
      & $s_r$ & $s_i$ & $s_m$ & &       & $s_r$ & $s_i$ & $s_m$ \\ \cline{1-4} \cline{6-9}
$c_g$ & 0.630 & 0.895 & 0.817 & & $c_g$ & 0.698 & 0.857 & 0.845 \\
$c_s$ & 0.631 & 0.527 & 0.578 & & $c_s$ & 0.853 & 0.847 & 0.851 \\
$c_d$ & 0.260 & 0.421 & 0.000 & & $c_d$ & 0.817 & 0.845 & 0.845 \\ \cline{1-4} \cline{6-9}
\\
\multicolumn{4}{c}{YaleB ($\bar{n} \approx 64$)} & &
\multicolumn{4}{c}{YTF ($\bar{n} \approx 251$)} \\
      & $s_r$ & $s_i$ & $s_m$ & &       & $s_r$ & $s_i$ & $s_m$ \\ \cline{1-4} \cline{6-9}
$c_g$ & 0.888 & 0.900 & 0.892 & & $c_g$ & 0.790 & 0.791 & 0.791 \\
$c_s$ & 0.916 & 0.905 & 0.891 & & $c_s$ & 0.791 & 0.790 & 0.790 \\
$c_d$ & 0.809 & 0.886 & 0.889 & & $c_d$ & 0.791 & 0.791 & 0.791 \\ \cline{1-4} \cline{6-9}
\end{tabular}
\end{center}
\end{table}

It shows that $c_s$ usually leads to bad results, expected that $s_r$ is used.
The main reason is that selecting randomly slows down the process of generating tiny components, while other methods tend to remove vertices at junctions and vertices in the core areas are divided into pieces very fast.

With the increase of data volume, the differences between them become more and more significant.
On MNIST, that each cluster contains $7,000$ digits, $s_r$, $c_s$ and $c_d$ are poorly effective, while all measures are almost equally effective on YaleB and YTF.

\subsection{Robustness}
The robustness to selection rate $r$, connection size $t$, and $k$-nn graph size $k$ is analyzed.
We vary them in ranges $r \in \{0.1, 0.2, \cdots, 0.9\}$ and $t, k \in \{4, 6, \cdots, 30\}$ respectively.
Other settings are the same as \ref{sec:exp:setting}.

\begin{figure}
\centerline{\includegraphics[width=0.5\textwidth]{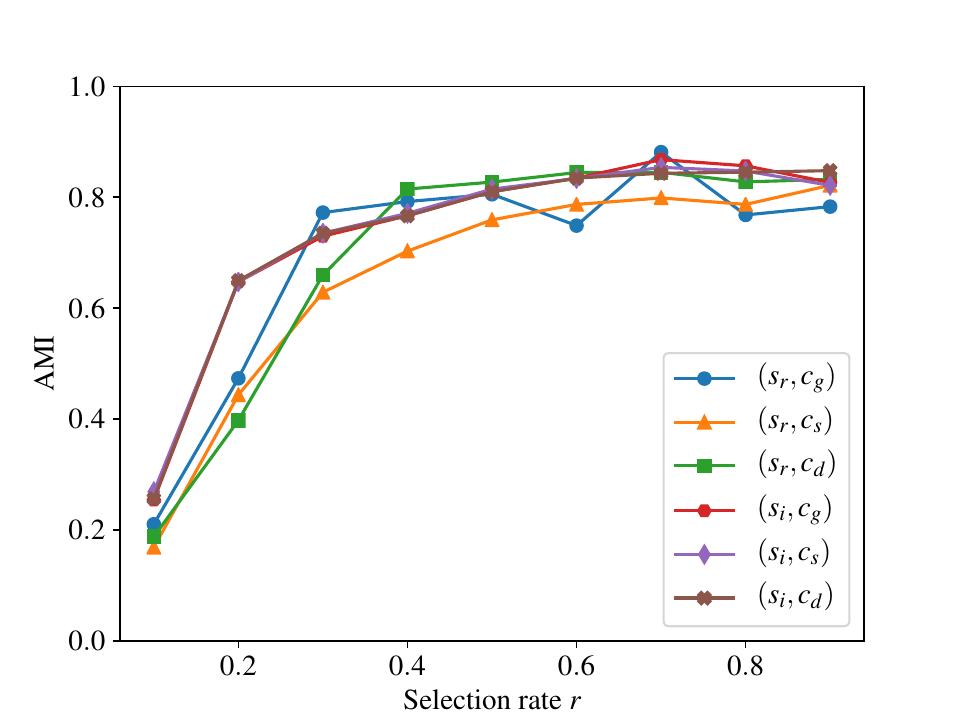}}
\caption{Robustness to selection rate $r$.
    It shows that the results are almost unchanged when $r \in (0.5, 1)$.
    The methods based on $s_i$ still preform very well, even $r$ has decreased to $0.2$.
}
\label{fig:rob-r}
\end{figure}

\begin{figure}
\centerline{\includegraphics[width=0.5\textwidth]{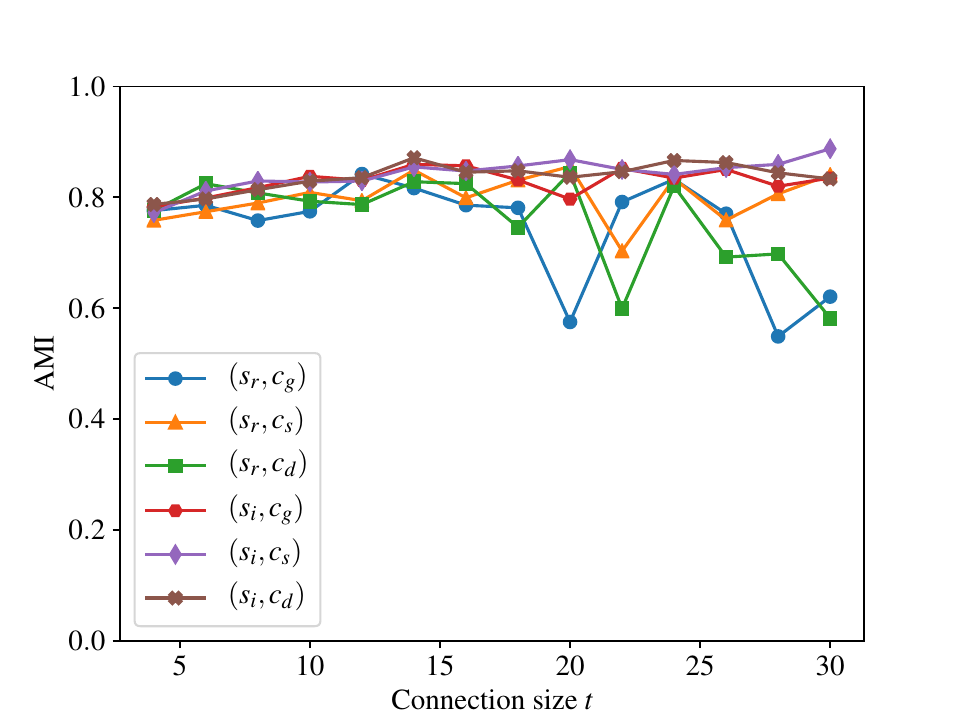}}
\caption{Robustness to connection size $t$.
    It shows that the method is quite stable when $t$ varies widely.
    However, it is worth noting that, since random selection $s_r$ is less efficient at separating connected clusters than $s_i$, when $t$ is very large, that is, the graph is connected more tightly, the performance is significantly worse than others.
}
\label{fig:rob-t}
\end{figure}

\begin{figure}
\centerline{\includegraphics[width=0.5\textwidth]{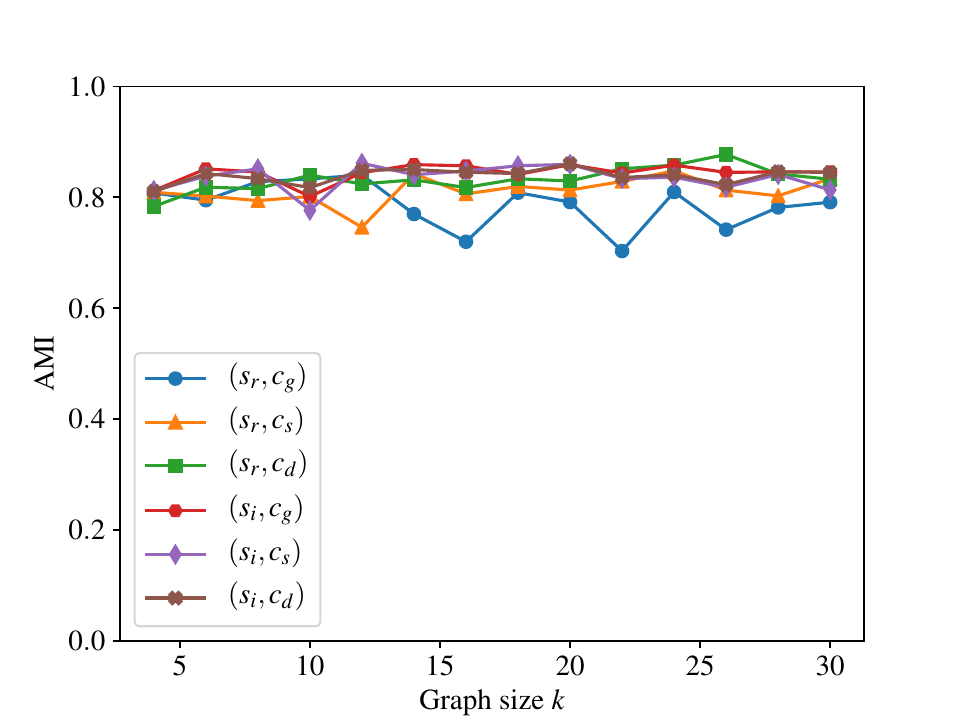}}
\caption{Robustness to graph size $k$.
    It clearly shows that all methods perform very well when $k$ varies in range $[4, 30]$, which means that the approach is not very strict with graph quality.
}
\label{fig:rob-k}
\end{figure}

Due to the limited space and the similarity between $s_i$ and $s_m$, we only consider six settings, $\{s_r, s_i\} \times \{c_g, c_s, c_d\}$, and evaluate them on Pendigits.
The results are shown in Fig. \ref{fig:rob-r}, \ref{fig:rob-t} and \ref{fig:rob-k} respectively.
Obviously, any parameter doesn't play a critical role for most measures in a wide range.

\subsection{Effectiveness of Smoothing}
\begin{figure}
\centerline{\includegraphics[width=0.5\textwidth]{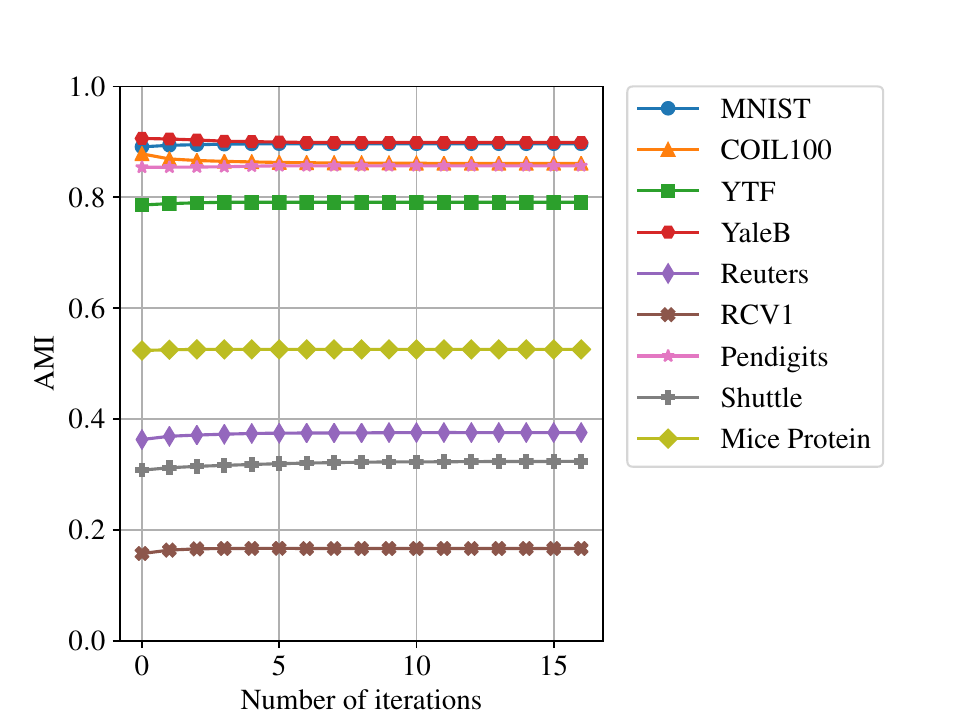}}
\caption{Effectiveness of smoothing.
    It shows that the fine-tuning method is effective on most datasets, especially for data with poor clustering quality.
    And besides, it is obvious that the method converges very fast.
}
\label{fig:smooth}
\end{figure}

We also test the effectiveness of smoothing.
Instead of running 16 times before and after pruning as the previous experiments, to make the effect on partitions more obvious, we only run it once before pruning to remove tiny clusters and 16 times after that.
The results, Fig. \ref{fig:smooth}, shows that it is effective on most datasets and converges very fast.

\subsection{Visualization}
\begin{figure}
\centerline{\includegraphics[width=0.5\textwidth]{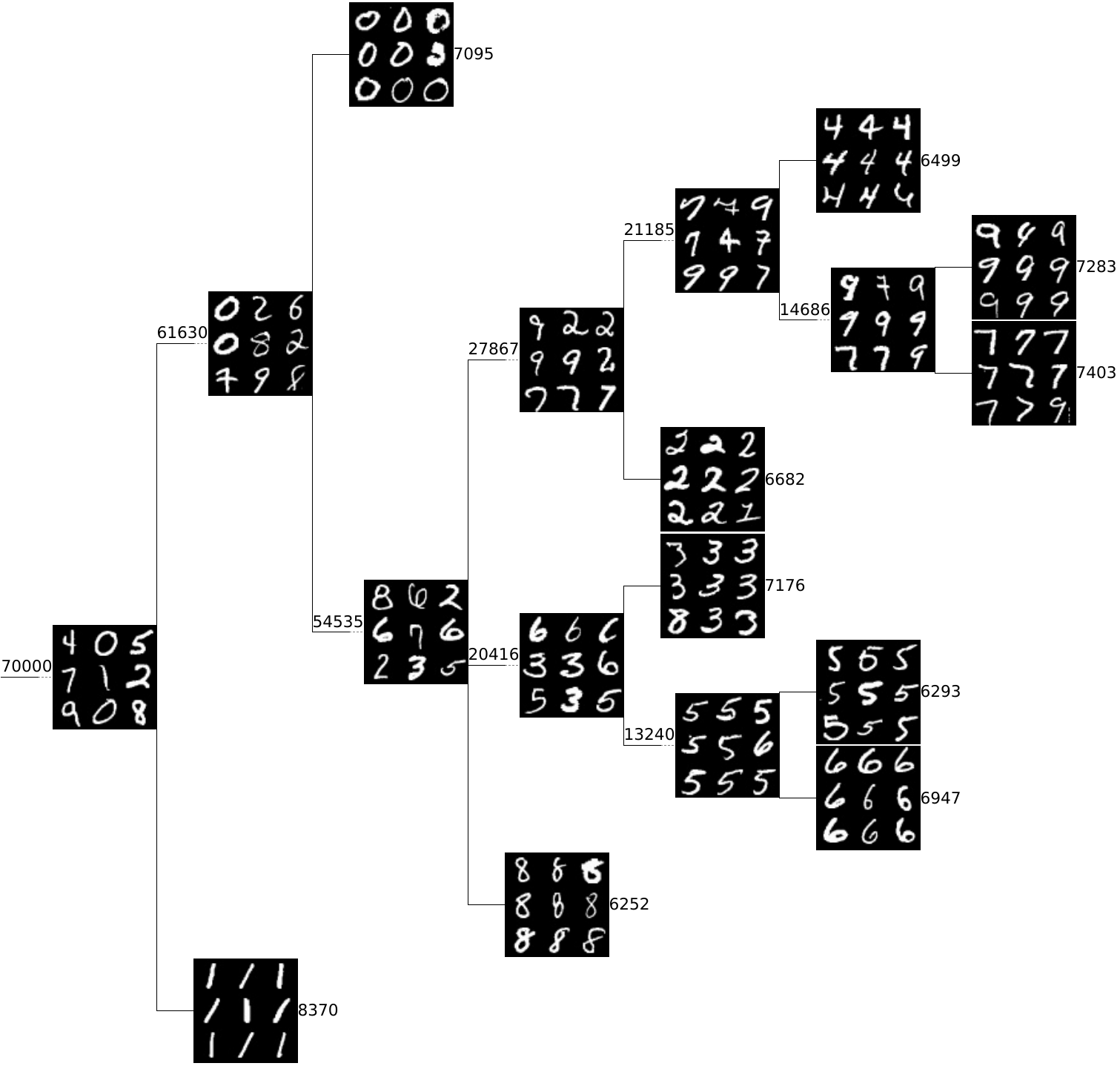}}
\caption{Visualization of the dendrogram on MNIST.}
\label{fig:visual-merge}
\end{figure}

The visualization of the result on MNIST is shown in Fig. \ref{fig:visual-merge}.
The number on a branch indicates the size of it, and the images are $9$ random samples in it.

The dendrogram shows that it first separates $1$s from the whole, and then $0$s.
The remaining digits are roughly divided into three groups, $\{8\}$, $\{3, \{5, 6\}\}$ and $\{2, \{4, \{7, 9\}\}\}$.

\section{Discussion on Experimental Results}\label{sec:discuss}
As shown in Table \ref{tab:AMI}, \textit{Reductive Clustering} (RC) is compared with 12 methods, and achieves the best results on 3 out of 9 datasets.
It is also the second fastest method on RCV1, and just a little slower, about $4\%$, than the fastest one.
The memory required is also less than all other algorithms except $k$-Means.
Moreover, it is guaranteed that the implementation runs in linear time.

Four of the datasets are images, including MNIST, COIL100, YTF and YaleB.
RC achieves the best result on the biggest dataset, MNIST, and the second best result on YTF.
On other two datasets, RC is only worse than RCCs and GDLs.
Additionally, it also one of the only two applicable hierarchical methods on MNIST.
Neither classical nor recent state-of-art agglomerative methods can handle such a large dataset.
Moreover, as mentioned earlier, GMM is not applicable on high-dimensional datasets.	

Reuters and RCV1 are two text datasets.
Unfortunately, almost all algorithms do not work well on them.
It is mainly because both of them are complicated in structure, and the measure, Euclidean distance of item frequency, is not effective and thus can not accurately represent the similarity.
For Shuttle and Mice Protein, there also exist similar problems.
The results on such datasets are greatly influenced by random factors.
It leads to that conservative methods like $k$-Means can achieve not bad results instead.
We believe that such results are not so reliable and should not be used as a key evaluation criteria like others.

\section{Conclusions}\label{sec:con}
We present an efficient graph-based divisive clustering approach which is shown to be effective on various types of data.
We also take an introductory discussion on strategies to divide faster and to avoid undesirable divisions.
The selection and connection measures discussed are very basic, but they work very well already.
We believe that the performance of the approach can be further improved with the help of more effective measures, especially for data of specific types.

As far as we know, it is the first schema trying to group data into clusters through repeatedly reducing a graph into a concise one, and also one of the few effective hierarchical methods running in linear time.
Existing methods may be adapted based on the schema, and the performances can be greatly improved.
We hope this work can inspire further research.

In addition to clustering algorithms, the effectiveness of cluster analysis is also heavily influenced by many other factors.

Preprocessing is the basis.
The experimental results show that most algorithms can not handle datasets like Reuters directly.
One of the main reasons is that the similarity measure used is not effective, which leads to the poor quality of graphs.

For hierarchical methods, post-processing is also important.
In many cases, a dendrogram may need to be pruned or converted into partitions.
Although it is independent of clustering approaches, considering that most existing pruning methods, like depth-based ones, are not so effective on it, we introduce two simple algorithms in this paper to make the clustering process complete.
However, as future work, it is obvious that, since they are simply based on the sizes of leaves, they can be further optimized.

\bibliographystyle{IEEEtran}
\bibliography{rc}
\end{document}